\documentclass{article}
\PassOptionsToPackage{numbers}{natbib}

\usepackage[main, final]{neurips_2026}

\usepackage[utf8]{inputenc} 
\usepackage[T1]{fontenc}    
\usepackage{hyperref}       
\usepackage{url}            
\usepackage{booktabs}       
\usepackage{amsfonts}       
\usepackage{nicefrac}       
\usepackage{microtype}      
\usepackage{xcolor}         

\usepackage{microtype}
\usepackage{hhline}
\usepackage{makecell}
\usepackage{multirow}
\usepackage{xcolor}
\definecolor{lightred}{rgb}{1, 0.2, 0.2}
\definecolor{lightgreen}{rgb}{0.25, 0.75, 0.25}
\definecolor{iccvblue}{rgb}{0.26,0.39,0.85}
\usepackage{amssymb}
\usepackage{tabularx}
\usepackage{soul}
\usepackage{microtype}
\usepackage{graphicx}
\usepackage{booktabs}
\usepackage{colortbl}
\usepackage{pifont}
\usepackage{enumitem}
\usepackage{amsmath}
\usepackage{hyperref}

\hypersetup{
    colorlinks=true,      
    linkcolor=iccvblue, 
    citecolor=iccvblue, 
    urlcolor=iccvblue
}

\title{3R-Adapter: Retrieval, Rewiring, and Refinement for Efficient Adaptation of 3D Reconstruction Model}

\author{%
  Zixiang Ai$^{1}$ \quad
  Zhenyu Cui$^{2}$ \quad
  Xu Zou$^{3}$ \quad
  Jiahuan Zhou$^{1,\dagger}$\\
  \normalfont
  $^{1}$Peking University \enspace
  $^{2}$Tsinghua University \enspace
  $^{3}$Huazhong University of Science and Technology
}

\begin{document}

\maketitle

\renewcommand{\footnoterule}{}
\renewcommand{\thefootnote}{\fnsymbol{footnote}}
\footnotetext[2]{Corresponding author: \texttt{jiahuanzhou@pku.edu.cn}}

\begin{abstract}\label{abs}
Feed-forward 3D reconstruction models have demonstrated strong generalization under large-scale pre-training, yet they struggle in long-tail scenarios such as low-light reconstruction, joint human-scene reconstruction, and test-time adaptation. 
We observe relational attention structure collapse, manifested as the failure of cross-view correspondences, as a primary cause of these limitations.
While adaptation offers a practical way to bridge this gap, existing parameter-efficient fine-tuning (PEFT) approaches mainly perform channel-wise feature adjustment and lack mechanisms to explicitly adapt relational structures, leaving correspondence failures unresolved.
In this work, we formulate adaptation of 3D reconstruction models as a problem of correspondence recovery under domain shift and propose 3R-Adapter, a PEFT paradigm for long-tail 3D reconstruction scenarios. Our method consists of three components: Deformable Retrieval for task-adaptive relation candidate retrieval, Relational Rewiring for reconstructing cross-view relational structures, and Consistency Refinement for injecting the rewired relations into stable predictions.
Extensive experiments on low-light reconstruction, joint human-scene reconstruction, and test-time adaptation demonstrate that our approach significantly improves robustness over existing PEFT methods while maintaining efficient lightweight adaptation. Our code is released at \texttt{https://github.com/PKU-OV3-LAB/3R-Adapter}.
\end{abstract}

\section{Introduction}\label{intro}
\begin{figure}[!t]
\centering
\includegraphics[width=\textwidth]{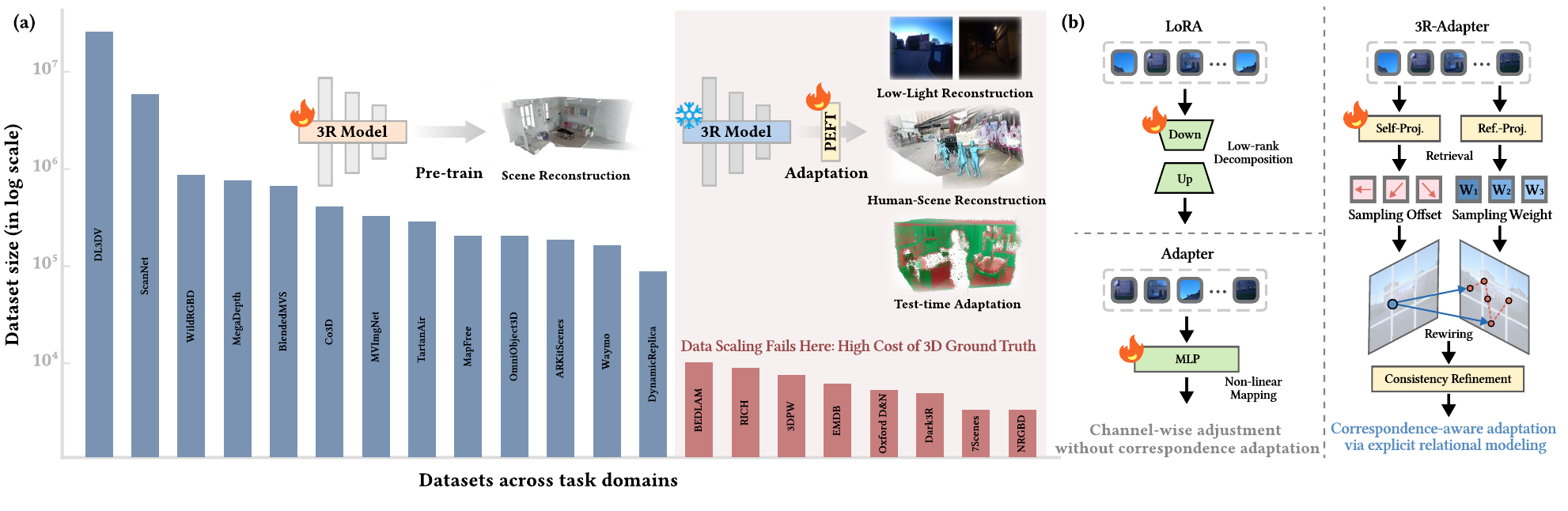}
\caption{\textbf{The data scaling bottleneck in 3D pre-training and our relational adaptation paradigm.} (a) While foundation models thrive on massive data, they struggle to generalize to long-tail 3D domains due to the high cost of dense 3D ground truth. (b) Conventional PEFT methods (e.g., LoRA, Adapter) rely on implicit, channel-wise feature adjustment, which struggle to model the correspondence relationship in data-scarce scenarios. In contrast, our 3R-Adapter shifts the paradigm to explicit relational adaptation via Deformable Retrieval, Relational Rewiring, and Consistency Refinement, effectively establishing robust cross-view correspondences without destroying pre-trained priors.}
\label{fig-motivation}
\end{figure}

Recent feed-forward 3D reconstruction models~\cite{dust3r,cut3r,vggt,pi3,da3} have demonstrated strong generalization ability under large-scale pre-training. These models leverage large-scale data~\cite{dl3dv,scannet,wildrgbd,megadepth,co3d,mvimgnet,tartanair} to implicitly learn geometric structures through attention-based representations, enabling the estimation of camera poses, intrinsics, and dense geometry from uncalibrated image collections. 
Despite these advances, current models exhibit clear limitations under domain shift. In scenarios such as low-light environments~\cite{dark3r}, joint human-scene reconstruction~\cite{human3r,unish,josh3r}, and test-time adaptation~\cite{test3r,online3r}, data distributions deviate significantly from pre-training conditions and reliable 3D supervision is often unavailable, as depicted in Fig.~\ref{fig-motivation}. As a result, pre-trained models frequently fail to generalize, leading to unstable geometric predictions.

\begin{figure}[!t]
\centering
\includegraphics[width=\textwidth]{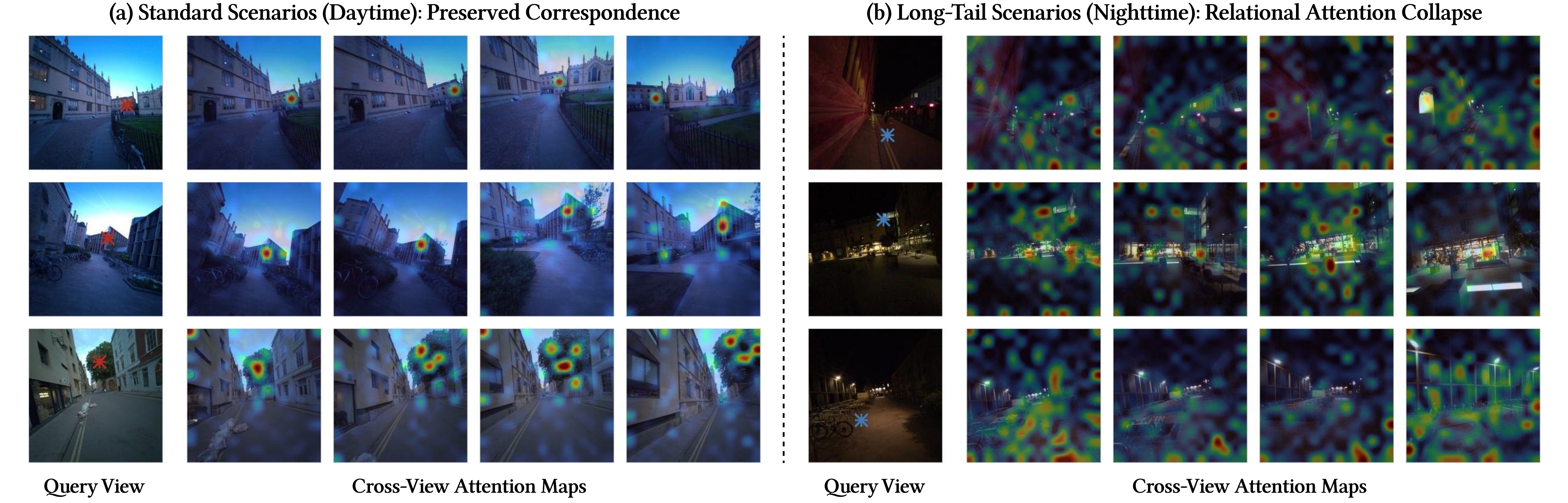}
\caption{
\textbf{Relational attention structure collapse in long-tail scenarios.}
(a) In data-rich daytime scenes, the pre-trained 3D reconstruction model produces concentrated cross-view attention, allowing a selected query token to attend to consistent physical counterparts across reference views.
(b) For selected low-light queries, the attention distribution becomes diffuse and fragmented, indicating degraded relational routing and unreliable correspondence modeling.
This observation suggests that long-tail adaptation requires explicitly restoring cross-view relational structures, motivating the design of 3R-Adapter.
}
\label{fig-observation}
\end{figure}

We observe that this failure is closely related to the instability of cross-view relational modeling, as shown in Fig.~\ref{fig-observation}. Specifically, attention patterns that are well-structured in in-domain settings become diffuse and less informative under domain shift, making it difficult to establish reliable cross-view correspondences. This suggests that the core challenge lies not in insufficient feature representation, but in the inability to preserve and adapt relational structures across views.

Addressing this issue requires effective adaptation mechanisms that can restore stable cross-view relations. Parameter-efficient fine-tuning (PEFT)~\cite{vpt,dept,gaprompt,ai2025vgp,ai2026gaprompt++} provides a practical solution by enabling lightweight adaptation of large pre-trained models. However, existing PEFT methods, such as LoRA~\cite{lora} and Adapter~\cite{adapter}, primarily perform channel-wise feature modulation and lack mechanisms to explicitly adapt relational structures. Consequently, they rely on implicit adjustments of attention to capture cross-view dependencies, offering limited control over relational routing and often failing to recover stable correspondences under domain shift, demonstrated in Fig.~\ref{fig-motivation}.

In this work, we argue that effective adaptation for feed-forward 3D reconstruction requires relational structure recovery. We therefore reformulate model adaptation as a problem of relational adaptation rather than feature reparameterization. Under this perspective, PEFT is not merely a tool for adjusting feature distributions, but a mechanism for introducing task-level structural constraints that explicitly guide cross-view correspondence modeling.

Based on this formulation, we propose 3R-Adapter, a relation-centric PEFT framework for long-tail adaptation. The proposed method consists of three components. Deformable Retrieval, which performs task-adaptive relational querying by selecting appropriate anchor tokens and retrieving candidate cross-view features via deformable sampling. Relational Rewiring, which reconstructs cross-view relational structures by reorganizing retrieved candidates into consistent associations. Consistency Refinement, which injects reconstructed relations into the prediction space to enforce stable and consistent geometric outputs. 
Together, these modules enable efficient and explicit adaptation of relational structures while preserving the generalization capability of the pre-trained backbone.

In summary, our contributions are threefold:
\begin{itemize}
[leftmargin=1.5em]
\item{We identify relational structure collapse as a fundamental failure mode of feed-forward 3D reconstruction models under domain shift, and show that effective adaptation requires restoring cross-view relational structures rather than merely adjusting features. 
}
\item{
We propose 3R-Adapter, a relation-centric PEFT framework that enables explicit adaptation of cross-view associations through retrieval, rewiring, and refinement. To our knowledge, this is the first work that explores parameter-efficient adaptation specifically designed for feed-forward 3D reconstruction models.
}
\item{
Extensive experiments demonstrate that, with minimal parameter overhead, 3R-Adapter significantly improves robustness in long-tail scenarios, including joint human-scene reconstruction, low-light conditions, and test-time adaptation.
}
\end{itemize}

\section{Observation and Insight}
\label{observ}
Feed-forward 3D reconstruction models rely on attention-based relational structures to establish cross-view correspondences across input images~\cite{croco}. 
While such structures are effectively learned from data-rich pre-training scenarios, their behavior in data-sparse long-tail scenarios remains less reliable. 
To better understand this failure mode, we visualize cross-view attention patterns of pre-trained reconstruction models in standard daytime scenes and challenging low-light scenes.

As shown in Fig.~\ref{fig-observation}, in standard data-rich scenarios, the attention distribution of a selected query token is concentrated around consistent physical counterparts across reference views. 
This indicates that the pre-trained model can form stable relational routes when the input distribution is close to the pre-training regime. 
In contrast, for selected queries under extreme low-light conditions, the attention distribution becomes diffuse and fragmented, making it difficult to maintain reliable cross-view correspondences. 
We refer to the degradation of geometrically consistent cross-view associations as \textit{relational attention structure collapse}. The quantitative correspondence analysis in Sec.~\ref{sec-correspondence-accuracy} measures this geometric degradation and its recovery after adaptation.

This observation suggests that the degradation of feed-forward 3D reconstruction models in long-tail scenarios is not merely caused by weakened local feature representations. 
A key bottleneck lies in the instability of cross-view relational structures that support correspondence modeling. 
Therefore, effective adaptation should not only adjust channel-wise feature distributions, but also explicitly restore and reorganize cross-view relations. 
This motivates our 3R-Adapter, which retrieves task-relevant relational candidates, rewires them into structured cross-view associations, and refines predictions through consistency-preserving residual updates.

\section{Related Work}
\label{related}
\subsection{3D Reconstruction Model}
Traditional 3D reconstruction methods like structure-from-motion (SfM)~\cite{sfm} and multi-view stereo (MVS)~\cite{mvs} rely on establishing cross-view correspondences through handcrafted or learned local features, followed by geometrically grounded optimization with strong 3D inductive priors. These pipelines remain sensitive to weak texture, limited overlap, and viewpoint changes, and are computationally costly.

Feed-forward 3D reconstruction models~\cite{dust3r,mast3r,spann3r,cut3r,vggt,da3,pi3} provide a promising alternative by learning geometric reasoning directly from data. Pioneering work DUSt3R~\cite{dust3r} reformulates reconstruction as dense pointmap prediction from uncalibrated image pairs. CUT3R~\cite{cut3r} introduces a persistent-state recurrent formulation, allowing the model to continuously update a shared 3D state from streaming observations and produce metric-scale pointmaps online. VGGT~\cite{vggt} further scales feed-forward reconstruction to hundreds of views, directly predicting camera parameters, depth maps, point maps, and point tracks with a unified transformer architecture.  More recent approaches such as $\pi^3$~\cite{pi3} remove the reliance on a fixed reference view by adopting permutation-equivariant visual geometry learning, while Depth Anything 3~\cite{da3} investigates a minimal formulation based on a single transformer and depth-ray representation for joint any-view depth and pose estimation. These approaches demonstrate that correspondence and geometry can be implicitly learned at scale, marking a shift from explicit optimization to data-driven geometric reasoning.

Despite this progress, existing feed-forward models are primarily optimized for in-distribution performance and exhibit limited robustness under domain shift. In scenarios such as low-light conditions, the learned cross-view correspondences often become unreliable. This highlights the need for effective adaptation mechanisms that can recover stable cross-view relations under distribution shifts.

\subsection{Parameter-efficient Fine-tuning}
With the increasing scale of models, full fine-tuning has become computationally prohibitive for downstream adaptation. To address this challenge, parameter-efficient fine-tuning (PEFT) methods~\cite{vpt,lora,adapter} have been widely explored in both vision and language domains.

Existing PEFT approaches can be broadly categorized into three families. Prompt-based methods, such as visual prompt tuning~\cite{vpt}, introduce a small set of learnable tokens to steer pretrained models without modifying backbone parameters. Adapter-based methods~\cite{adapter,nightadapter} insert lightweight bottleneck modules into intermediate layers to refine latent feature representations. Reparameterization methods, including LoRA~\cite{lora} and its variants such as DoRA~\cite{dora}, approximate weight updates via low-rank decompositions, enabling efficient adaptation with minimal trainable parameters.
Recent works have extended these paradigms to feed-forward 3D reconstruction. For example, Test3R~\cite{test3r} and Online3R~\cite{online3r} employ prompt-based adaptation with self-supervised objectives for test-time adaptation on single scenes, while Gen3R~\cite{gen3r} introduces lightweight adapters to bridge representation gaps across models. Meanwhile, methods such as VIST3A~\cite{vist3a} and VGGRPO~\cite{vggrpo} adopt LoRA-style reparameterization to enable efficient alignment and adaptation of large geometry models. These works demonstrate that PEFT  techniques provide a practical solution for adapting feed-forward reconstruction models under limited supervision.

However, despite their effectiveness, existing methods primarily operate through feature-level modulation and rely on implicit updates of attention to capture cross-view dependencies. In multi-view 3D reconstruction, where geometric reasoning critically depends on structured cross-view relations, such implicit adaptation provides limited control over relational routing and often fails to maintain stable correspondences under domain shift. This limitation motivates our work, where we explicitly model and adapt cross-view relational structures through a relation-centric PEFT framework.




\section{Method}
\label{method}
\begin{figure}[!t]
\centering
\includegraphics[width=\textwidth]{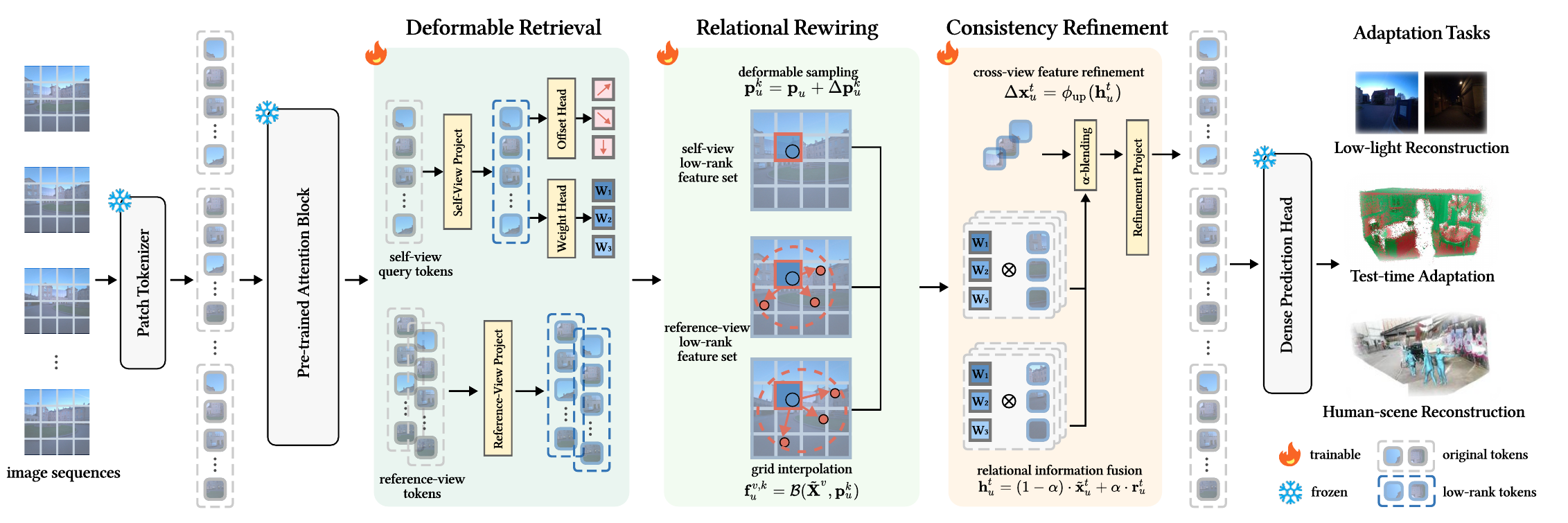}

\caption{\textbf{The overall pipeline of the proposed 3R-Adapter.} To enable efficient adaptation for complex long-tail scenes, our 3R-Adapter modulates the network via three cohesive modules: Deformable Retrieval extracts view-specific anchors and estimates spatial offsets alongside routing weights; Relational Rewiring establishes explicit geometric correspondences across views through deformable spatial sampling; and Consistency Refinement integrates the relational features back into the main stream via a learned gating mechanism, effectively enforcing cross-view consistency.}
\label{fig-pipeline}
\end{figure}

In this section, we detail our method for efficiently adapting the pre-trained 3D reconstruction models, improving the correspondence capturing, thus enhancing the reconstruction quality. To explicitly guide cross-view correspondence modeling beyond feature shift in original pre-trained models, 
we introduce 3R-Adapter with three modules: Deformable Retrieval, Relational Rewiring, and Consistency Refinement, demonstrated in Fig.~\ref{fig-pipeline}. Given a pre-trained model, we freeze the parameters and only update our newly inserted modules.

\subsection{Preliminary: Feed-forward 3D Reconstruction}
Given a set of input images $\mathcal{I} = \{I_i\}_{i=1}^N$, feed-forward 3D reconstruction models first encode them into patch-level feature tokens. We denote the flattened token set across all views as $\mathbf{X} = \{\mathbf{x}_k\}_{k=1}^M \in \mathbb{R}^{M \times C}$, where $M = N \times H \times W$ is the total number of tokens. The core mechanism of these models is to establish cross-view geometric correspondences by learning a relational function $\mathcal{R}_\theta$, yielding contextualized representations:
\begin{equation}
    \mathbf{Z} = \mathcal{R}_\theta(\mathbf{X}).
\end{equation}

Recent vision foundation models typically instantiate $\mathcal{R}_\theta$ through two representative architectural paradigms:

\textbf{Global Attention-based Models.} Architectures such as VGGT~\cite{vggt} and $\pi^3$~\cite{pi3} capture cross-view interactions via global attention. For a given token $\mathbf{x}_i$, its relational output $\mathbf{z}_i$ is computed by attending to all context tokens:
\begin{equation}
    \mathbf{z}_i = \sum_j \text{Softmax}(\mathbf{Q}_i \mathbf{K}_j^\top)\mathbf{V}_j,
\end{equation}
where $j \in \{1, \dots, M\}$.

\textbf{Recurrent State-based Models.} Alternatively, models such as CUT3R~\cite{cut3r} maintain a shared latent state to aggregate multi-view information sequentially:
\begin{equation}
    \mathbf{S}_t = g_\theta(\mathbf{S}_{t-1}, \mathbf{X}_t),
\end{equation}
where $\mathbf{S}_t$ encodes accumulated cross-view relations up to time step $t$.
Despite their architectural differences, both paradigms rely on large-scale data to implicitly learn the relational structure $\mathcal{R}_\theta$. Under domain shift in long-tail scenarios, this learned structure often becomes unstable, leading to degraded correspondence estimation. This limitation motivates the need for explicit relational adaptation.

\subsection{Deformable Retrieval}

A straightforward way to adapt this relation function is to recompute dense interactions among all tokens across all views. 
However, for $N$ views with feature resolution $H \times W$, dense all-to-all matching introduces quadratic interactions and can also amplify noisy correlations under domain shift. 
To this end, we design a lightweight Deformable Retrieval module to query task-relevant relational candidates. 

Firstly, we restrict the relational search space by introducing a structured sliding window over the view dimension. 
For a target view $t$, tokens from $\mathbf{X}^{t} = \{\mathbf{x}^{t}_{u}\}_{u=1}^{HW} \in \mathbb{R}^{HW \times C}$ serve as self-view query tokens, while tokens from neighboring views $\mathbf{X}^{v}$ are used as reference tokens. 
Specifically, we restrict reference views by a sliding window over the view dimension:
\begin{equation}
    \mathcal{N}(t) = \{v \mid |v - t| \leq w \},
\end{equation}
where $w$ is the window radius. 
This reduces the number of view-level interactions from $O(N^2)$ to $O(Nw)$ and provides a coarse locality prior over the relational graph.

For each self-view query token $\mathbf{x}^{t}_{u}$ with spatial position $\mathbf{p}_{u}$, we first project it into a low-dimensional latent space:
\begin{equation}
    \tilde{\mathbf{x}}^{t}_{u} = \phi_{self}(\mathbf{x}^{t}_{u}) \in \mathbb{R}^{r},
\end{equation}
where $r \ll C$. 
From this compact query representation, we predict $K$ deformable offsets and their corresponding sampling weights:
\begin{equation}
    \{\Delta \mathbf{\hat{p}}_{u}^{k}\}_{k=1}^{K} = \psi_{\Delta}(\tilde{\mathbf{x}}^{t}_{u}), 
    \quad
    \{\mathbf{w}_{u}^{k}\}_{k=1}^{K} = \psi_{w}(\tilde{\mathbf{x}}^{t}_{u}).
    \label{eq1}
\end{equation}

In parallel, reference tokens are projected into the same low-rank space:
\begin{equation}
    \tilde{\mathbf{X}}^{v} = \phi_{ref}(\mathbf{X}^{v}) \in \mathbb{R}^{HW \times r}, 
    \quad v \in \mathcal{N}(t).
\end{equation}
Thus, Deformable Retrieval defines a sparse set of candidate relation locations $\{\mathbf{p}_{u}^{k}\}$ and weights $\{\mathbf{w}_{u}^{k}\}$ for each self-view token, while keeping both query and reference features in a lightweight latent space.

Compared with dense attention, Deformable Retrieval introduces only a small number of trainable parameters and provides a structured search space for relational adaptation, making it well-suited for parameter-efficient fine-tuning without disrupting the pre-trained backbone. 
The sliding window constrains the view-level relational graph, while the deformable offsets provide fine-grained spatial flexibility within selected reference views. 
The resulting sparse candidates are then materialized and aggregated in the following Relational Rewiring stage.

\subsection{Relational Rewiring}

Given the offsets and weights predicted by Deformable Retrieval, Relational Rewiring explicitly constructs sparse cross-view relations through deformable sampling and weighted aggregation. 
For each self-view query token $\mathbf{x}^{t}_{u}$ with spatial position $\mathbf{p}_{u}$, we first obtain $K$ candidate sampling locations:
\begin{equation}
    \mathbf{p}_{u}^{k} = \mathbf{p}_{u} + \Delta \mathbf{p}_{u}^{k},
\end{equation}
where $\Delta \mathbf{p}_{u}^{k}$ is the $k$-th predicted offset.

With the predicted raw offsets $\{\Delta \mathbf{\hat{p}}_{u}^{k}\}_{k=1}^{K}$ in Equation \ref{eq1}, we constrain the search range by applying a bounded transformation:
\begin{equation}
    \Delta \mathbf{p}_{u}^{k}
    =
    \rho \cdot \tanh(\Delta \mathbf{\hat{p}}_{u}^{k}),
\end{equation}
where $\rho$ is a radius hyper-parameter controlling the maximum deformable search range. 
The candidate sampling location is then computed as:
\begin{equation}
    \mathbf{p}_{u}^{k}
    =
    \mathbf{p}_{u} + \Delta \mathbf{p}_{u}^{k}.
\end{equation}

After that, we reshape the projected reference feature map $\tilde{\mathbf{X}}^{v} \in \mathbb{R}^{HW \times r}$ into a spatial feature grid $\tilde{\mathbf{X}}^{v} \in \mathbb{R}^{H \times W \times r}$ and view it as a continuous feature field via bilinear interpolation, where $v \in \mathcal{N}(t)$
The candidate relational feature is retrieved by:
\begin{equation}
    \mathbf{f}_{u}^{v,k}
    =
    \mathcal{B}(\tilde{\mathbf{X}}^{v}, \mathbf{p}_{u}^{k}),
\end{equation}
where $\mathcal{B}(\cdot,\cdot)$ denotes bilinear sampling on the reference feature grid.
The sampled candidates are then aggregated into a rewired relational representation:
\begin{equation}
    \mathbf{r}_{u}^{t} = \sum_{v \in \mathcal{N}(t)} \sum_{k=1}^{K} \mathbf{w}_{u}^{k}\, \mathbf{f}_{u}^{v,k},
\end{equation}
where $a_{u}^{k}$ denotes the learned importance weight for the $k$-th candidate and $\mathbf{r}_{u}^{t} \in \mathbb{R}^{r}$ is the relation-enhanced representation for query token $\mathbf{x}^{t}_{u}$.

This process converts the offsets and weights predicted in Deformable Retrieval into explicit cross-view associations. 
Instead of relying on unconstrained all-to-all attention, Relational Rewiring constructs a sparse and task-adaptive relational graph, enabling more stable correspondence modeling under domain shift.

\subsection{Consistency Refinement}

After Relational Rewiring, each self-view query token $\mathbf{x}^{t}_{u}$ is associated with a rewired relational representation $\mathbf{r}^{t}_{u} \in \mathbb{R}^{r}$. 
Rather than directly modifying the pre-trained backbone features, we inject this relational information through a lightweight residual branch, which preserves the original representation while enabling task-specific adaptation.

Specifically, we fuse the self-view latent token $\tilde{\mathbf{x}}^{t}_{u}$ and the rewired relational representation $\mathbf{r}^{t}_{u}$ in the adapter space:
\begin{equation}
\mathbf{h}^{t}_{u}=(1-\alpha)\cdot\tilde{\mathbf{x}}^{t}_{u}+\alpha \cdot \mathbf{r}^{t}_{u},
\end{equation}
where $\alpha$ controls the strength of relational blending. 
The fused representation is then projected back to the original feature dimension:
\begin{equation}
\Delta \mathbf{x}^{t}_{u}=\phi_{\mathrm{up}}(\mathbf{h}^{t}_{u}) \in \mathbb{R}^{C}.
\end{equation}

Finally, the adapter output is added to the frozen backbone feature through a residual connection:
\begin{equation}
    \hat{\mathbf{x}}^{t}_{u}
    =
    \mathbf{x}^{t}_{u}
    +
    \gamma \Delta \mathbf{x}^{t}_{u},
\end{equation}
where $\gamma$ denotes the scaling factor. 
This residual formulation constrains the adaptation to be a lightweight correction to the pre-trained representation, reducing the risk of disrupting the original geometric priors while allowing the rewired cross-view relations to refine downstream predictions.

\section{Experiments}
\label{expe}
\subsection{Adaptation Tasks}

We evaluate 3R-Adapter on three representative adaptation settings: low-light reconstruction, joint human-scene reconstruction, and test-time adaptation. These tasks cover diverse forms of domain shift and long-tail scenarios where feed-forward 3D reconstruction models struggle to maintain stable cross-view relations.

\textbf{Low-light reconstruction.}
Under low-light conditions, severe image degradation and distribution shifts from pre-training data lead to unreliable feature extraction. Consequently, cross-view correspondences become unstable, making accurate geometric reconstruction challenging.

\textbf{Joint human-scene reconstruction.}
This task requires reconstructing both parametric human geometry (e.g., SMPL~\cite{smpl}) and background scene structure within a unified framework. The presence of non-rigid motion and cross-entity interactions introduces significant complexity, making it difficult for pre-trained models to recover consistent geometry and fine-grained human meshes.

\textbf{Test-time adaptation.}
In this setting, models must adapt to a single scene without additional supervision, relying on self-supervised signals or internal consistency at test time. This scenario is particularly sensitive to errors in cross-view relations, as no external supervision is available to correct them.

Overall, these tasks share a common challenge: maintaining stable cross-view correspondences under domain shift, making them suitable benchmarks for evaluating relational adaptation.

\subsection{Comparing Methods}

Our 3R-Adapter is evaluated across multiple feed-forward 3D reconstruction backbones, including DUSt3R~\cite{dust3r}, CUT3R~\cite{cut3r}, and $\pi^3$~\cite{pi3}, to demonstrate its generality and backbone-agnostic design.

We compare against three categories of methods. First, we include recent feed-forward reconstruction models across all tasks. Second, we consider task-specific approaches tailored to individual settings. For low-light reconstruction, we compare with Dark3R~\cite{dark3r}, which leverages contrastive learning between low-light and normal images. For joint human-scene reconstruction, we include Josh3R~\cite{josh3r}, Human3R~\cite{human3r}, and UniSH~\cite{unish}, which incorporate specialized designs for modeling human-scene interactions. For test-time adaptation, we compare with Test3R~\cite{test3r} and Online3R~\cite{online3r}, which refine predictions via self-supervised objectives on single scenes. 
Finally, we compare with representative parameter-efficient fine-tuning (PEFT) methods from NLP~\cite{dora,hydralora}, 2D vision~\cite{nightadapter,tlora}, and 3D vision~\cite{gaprompt}. These baselines share the same training protocols and evaluation settings as ours, enabling a fair comparison of adaptation efficiency and effectiveness.

\begin{table*}[!t]
\centering
\caption{
Comparison on the \textit{Oxford Day\&Night}~\cite{oxforddn} test dataset for camera pose estimation and multi-view scene reconstruction. 
}
\vspace{5pt}
\begin{small}
\resizebox{\textwidth}{!}{
\begin{tabular}{lcc cccc cccc}
\toprule
\multirow{2}{*}{\textbf{Method}} 
& \multirow{2}{*}{\textbf{Publication}}
& \multirow{2}{*}{\textbf{Params.}}
& \multicolumn{3}{c}{\textbf{Camera Pose Estimation}} 
& \multicolumn{4}{c}{\textbf{Scene Reconstruction}} \\

\cmidrule(lr){4-6} \cmidrule(lr){7-10}

& & 
& \textbf{~~~~ATE} $\downarrow$~~
& \textbf{~~RPE-T} $\downarrow$~~
& \textbf{~~RPE-R} $\downarrow$~~
& \textbf{Accuracy} $\downarrow$ 
& \textbf{Complete} $\downarrow$ 
& \textbf{Chamfer} $\downarrow$ 
& \textbf{Normal} $\uparrow$ \\

\midrule
DUSt3R~\cite{dust3r} & CVPR 24 & 0.57B 
& 0.937 & 0.729 & 10.351
& 0.741 & 0.598 & 1.339 & 0.526 \\

CUT3R~\cite{cut3r} & CVPR 25 & 0.79B 
& 0.784 & 0.542 & 9.924 
& 0.562 & 0.639 & 1.201 & 0.534 \\

VGGT~\cite{vggt} & CVPR 25 & 1.19B 
& 0.401 & 0.324 & 4.236
& 0.356 & 0.585 & 0.941 & 0.522 \\


Dark3R~\cite{dark3r} & CVPR 26 & 0.69B 
& 0.387 & 0.285 & 4.019
& 0.473 & 0.424 & 0.897 & 0.532 \\

$\pi^3$~\cite{pi3} & ICLR 26 & 0.96B 
& 0.345 & 0.261 & 3.657
& 0.190 & 0.746 & 0.936 & 0.518 \\

\midrule
\multicolumn{10}{c}{\textit{Parameter-efficient Fine-tuning on CUT3R}~\cite{cut3r}} \\
\midrule

DoRA~\cite{dora}          & ICML 24 & 5.6M
& 0.565 & 0.342 & 6.784
& 0.439 & 0.504 & 0.943 & 0.541 \\

HydraLoRA~\cite{hydralora}    & NeurIPS 24 & 10.0M 
& 0.612 & 0.457 & 7.719 
& 0.514 & 0.578 & 1.092 & 0.536 \\

GAPrompt~\cite{gaprompt}      & ICML 25 & 5.2M 
& 0.577 & 0.410 & 6.694 
& 0.467 & 0.484 & 0.951 & 0.532 \\

NightAdapter~\cite{nightadapter}  & ICCV 25 & 9.5M 
& 0.489 & 0.310 & 5.331 
& 0.375 & 0.416 & 0.791 & 0.544 \\

T-LoRA~\cite{tlora}  & AAAI 26 & 7.0M 
& 0.516 & 0.368 & 5.362 
& 0.350 & 0.489 & 0.839 & 0.538 \\

\rowcolor{gray!15}
\textbf{3R-Adapter} 
& This Paper
& \textbf{2.8M}
& \textbf{0.416} & \textbf{0.253} & \textbf{4.781}
& \textbf{0.254} & \textbf{0.327} & \textbf{0.581} & \textbf{0.550} \\

\midrule
\multicolumn{10}{c}{\textit{Parameter-efficient Fine-tuning on $\pi^3$}~\cite{pi3}} \\
\midrule

DoRA~\cite{dora}          & ICML 24 & 6.4M
& 0.376 & 0.245 & 3.459
& 0.198 & 0.543 & 0.741 & 0.524 \\

HydraLoRA~\cite{hydralora}    & NeurIPS 24 & 12.0M 
& 0.312 & 0.260 & 3.359 
& 0.178 & 0.624 & 0.802 & 0.526 \\

GAPrompt~\cite{gaprompt}      & ICML 25 & 6.2M 
& 0.396 & 0.291 & 3.852 
& 0.215 & 0.589 & 0.804 & 0.518 \\

NightAdapter~\cite{nightadapter}  & ICCV 25 & 10.3M 
& 0.297 & 0.225 & 3.341 
& 0.182 & 0.560 & 0.742 & 0.531 \\

T-LoRA~\cite{tlora}  & AAAI 26 & 7.9M 
& 0.326 & 0.269 & 3.340 
& 0.192 & 0.613 & 0.805 & 0.528 \\

\rowcolor{gray!15}
\textbf{3R-Adapter} 
& This Paper
& \textbf{3.2M}
& \textbf{0.258} & \textbf{0.199} & \textbf{3.199}
& \textbf{0.152} & \textbf{0.481} & \textbf{0.633} & \textbf{0.538} \\

\bottomrule
\end{tabular}
}
\end{small}
\label{tab-lowlight}
\end{table*}

\begin{table}[!ht]
\centering
\caption{Scene reconstruction under test-time adaptation (TTA) on \textit{7-Scenes}~\cite{7scene} and \textit{NRGBD}~\cite{nrgbd} datasets. Mean metrics are reported. Additional PEFT baselines are included for comparison.}
\begin{small}
\resizebox{1.0\textwidth}{!}{
\begin{tabular}{lcccccc|cccc}
\toprule
\multirow{2}{*}{\textbf{Method}} 
& \multirow{2}{*}{\textbf{Publication}} 
& \multirow{2}{*}{\textbf{Params.}}
& \multicolumn{4}{c}{\textbf{\textit{7-Scenes}}~\cite{7scene}} 
& \multicolumn{4}{c}{\textbf{\textit{NRGBD}}~\cite{nrgbd}} \\
\cmidrule(lr){4-7} \cmidrule(lr){8-11}
& & 
& \textbf{Accuracy} $\downarrow$ & \textbf{Complete} $\downarrow$ & \textbf{Chamfer} $\downarrow$ & \textbf{Normal} $\uparrow$
& \textbf{Accuracy} $\downarrow$ & \textbf{Complete} $\downarrow$ & \textbf{Chamfer} $\downarrow$ & \textbf{Normal} $\uparrow$ \\
\midrule
DUSt3R~\cite{dust3r}      & CVPR 24 & 0.57B & 0.146 & 0.181 & 0.327 & 0.736 
            & 0.144 & 0.154 & 0.299 & 0.871 \\
MASt3R~\cite{mast3r}      & CVPR 24 & 0.63B & 0.189 & 0.211 & 0.400 & 0.687 
            & 0.085 & 0.063 & 0.148 & 0.794 \\
CUT3R~\cite{cut3r}        & CVPR 25 & 0.79B & 0.126 & 0.154 & 0.280 & 0.727 
            & 0.099 & 0.076 & 0.175 & 0.837 \\
Test3R~\cite{test3r}      & NeurIPS 25 & 0.8M & 0.105 & 0.136 & 0.241 & 0.746 
            & 0.083 & 0.079 & 0.162& 0.870 \\
Online3R~\cite{online3r}  & CVPR 26  & 1.2M &0.112 & 0.145 & 0.257 & 0.733 
            & 0.084 & 0.081 & 0.165 & 0.829 \\
\midrule
\multicolumn{10}{c}{\textit{Parameter-Efficient Fine-Tuning on DUSt3R Backbone}} \\
\midrule
HydraLoRA~\cite{hydralora}   & NeurIPS 24 & 9.3M & 0.114 & 0.148 & 0.262 & 0.730 
            & 0.095 & 0.183 & 0.318 & 0.869 \\
GAPrompt~\cite{gaprompt}    & ICML 25 & 4.6M & 0.106 & 0.136 & 0.242 & 0.731 
            & 0.101 & 0.179 & 0.308 & 0.861 \\
NightAdapter~\cite{nightadapter}& ICCV 25 & 8.1M & 0.120 & 0.147 & 0.267 & 0.729
            & 0.093 & 0.142 & 0.235 & 0.865 \\
T-LoRA~\cite{tlora}      & AAAI 26 & 6.4M & 0.102 & 0.135 & 0.237 & 0.733
            & 0.089 & 0.131 & 0.220 & 0.872 \\

\rowcolor{gray!15}
\textbf{3R-Adapter} 
            & This Paper &  \textbf{2.5M}
            & \textbf{0.094} & \textbf{0.122} & \textbf{0.216} & \textbf{0.738} 
            & \textbf{0.082} & \textbf{0.123} & \textbf{0.205} & \textbf{0.879} \\
\bottomrule
\end{tabular}
}
\end{small}
\label{tab-tta}
\vspace{-15pt}
\end{table}

\subsection{Analysis on Low-light Reconstruction}

Table~\ref{tab-lowlight} reports results on the \textit{Oxford Day\&Night}~\cite{oxforddn} dataset. 
Low-light scenes severely degrade visual features and make cross-view matching unreliable, providing a challenging testbed for evaluating whether adaptation can recover stable relational structures.

3R-Adapter achieves the strongest PEFT performance on both CUT3R and $\pi^3$ backbones with the fewest trainable parameters. 
On CUT3R, it reduces Chamfer distance by about \textbf{27\%} compared with the strongest PEFT baseline, while using only \textbf{2.8M} trainable parameters. 
On the stronger $\pi^3$ backbone, 3R-Adapter also consistently improves pose estimation and reconstruction quality, reducing Chamfer distance by about \textbf{32\%} over the original backbone.

These gains indicate that the improvement does not come from increasing adaptation capacity, but from the proposed relation-centric design. 
By retrieving sparse relational candidates and rewiring them into structured cross-view associations, 3R-Adapter more effectively restores correspondence under low-light degradation than channel-wise feature adjustment.

\subsection{Analysis on Test-time Adaptation}

Test-time adaptation evaluates whether a model can refine reconstruction for each test scene without relying on additional supervised training data. 
This setting is particularly sensitive to unstable cross-view relations, since correspondence errors must be corrected using only scene-level self-supervised signals.

As shown in Table~\ref{tab-tta}, 3R-Adapter achieves the best PEFT performance on both \textit{7-Scenes} and \textit{NRGBD}. 
It reduces Chamfer distance by about \textbf{9\%} over the strongest PEFT baseline on \textit{7-Scenes} and by about \textbf{7\%} on \textit{NRGBD}, while using only \textbf{2.5M} trainable parameters. 
The consistent improvement across two datasets suggests that our method does not overfit to a single scene distribution.

Compared with specialized TTA methods such as Test3R~\cite{test3r} and Online3R~\cite{online3r}, 3R-Adapter remains competitive while following a lightweight relation-centric adaptation design. 
Instead of relying solely on self-supervised refinement at test time, our method explicitly retrieves and rewires reliable cross-view relations, providing a more structured way to improve scene-level consistency.

These results further support our central claim that effective adaptation for feed-forward 3D reconstruction should target relational structures, rather than only modulating channel-wise features.

\begin{table*}[!t]
\centering
\caption{
Comparison on the \textit{BEDLAM}~\cite{bedlam} validation dataset for joint human-scene reconstruction. 
}
\vspace{5pt}
\begin{small}
\resizebox{\textwidth}{!}{
\begin{tabular}{lcc ccc cccc}
\toprule
\multirow{2}{*}{\textbf{Method}} 
& \multirow{2}{*}{\textbf{Publication}}
& \multirow{2}{*}{\textbf{Params.}}
& \multicolumn{3}{c}{\textbf{Scene Reconstruction}} 
& \multicolumn{4}{c}{\textbf{Human Mesh Reconstruction}} \\

\cmidrule(lr){4-6} \cmidrule(lr){7-10}

& & 
& \textbf{Accuracy} $\downarrow$ 
& \textbf{Complete} $\downarrow$ 
& \textbf{Chamfer} $\downarrow$ 
& \textbf{PA-PVE} $\downarrow$  
& \textbf{Metric-PVE} $\downarrow$ 
& \textbf{PA-MPJPE} $\downarrow$ 
& \textbf{Metric-MPJPE} $\downarrow$ \\

\midrule

DUST3R~\cite{dust3r} & CVPR 24 & 0.57B & 0.237 & 0.159 & 0.396 & - & - & - & - \\
CUT3R~\cite{cut3r} & CVPR 25 & 0.79B & 0.185 & 0.096 & 0.281 & - & - & - & - \\
VGGT~\cite{vggt}   & CVPR 25 & 1.19B & 0.092 & 0.044 & 0.136 & - & - & - & - \\


\midrule
\multicolumn{10}{c}{\textit{Joint Human-Scene Reconstruction}} \\
\midrule
JOSH3R~\cite{josh3r} & ICLR 26 & 0.66B & 0.231 & 0.107 & 0.338 & 46.2 & 245.7 & 41.5 & 239.4 \\
Human3R~\cite{human3r} & ICLR 26 & 1.12B & 0.157 & 0.059 & 0.216 & 43.8 & 222.1 & 37.2 & 220.9 \\
UniSH~\cite{unish}  & CVPR 26 & 0.85B & 0.124 & 0.056 & 0.180 & 45.3 & 231.0 & 39.6 & 228.4 \\

\midrule
\multicolumn{10}{c}{\textit{Parameter-efficient Fine-tuning on CUT3R}~\cite{cut3r}} \\
\midrule
DoRA~\cite{dora}          & ICML 24 & 42.5M & 0.184 & 0.086 & 0.270 & 46.1 & 245.1 & 39.1 & 236.9 \\
HydraLoRA~\cite{hydralora}     & NeurIPS 24 & 48.0M & 0.201 & 0.124 & 0.325 & 47.8 & 258.8 & 40.5 & 251.4 \\
NightAdapter~\cite{nightadapter}  & ICCV 25 & 46.7M & 0.167 & 0.077 & 0.244 & 45.2 & 231.0 & 38.2 & 222.0 \\
T-LoRA~\cite{tlora}        & AAAI 26 & 40.3M & 0.159 & 0.068 & 0.227 & 45.5 & 227.2 & 37.9 & 225.4 \\
\rowcolor{gray!15}
\textbf{3R-Adapter} 
& This Paper
& \textbf{36.1M}
& \textbf{0.135} & \textbf{0.052} & \textbf{0.187}
& \textbf{44.4} & \textbf{218.5} & \textbf{37.6} & \textbf{217.2} \\

\bottomrule
\end{tabular}
}
\end{small}
\label{tab-humanscene}

\end{table*}

\begin{figure}[!t]
\centering
\includegraphics[width=\textwidth]{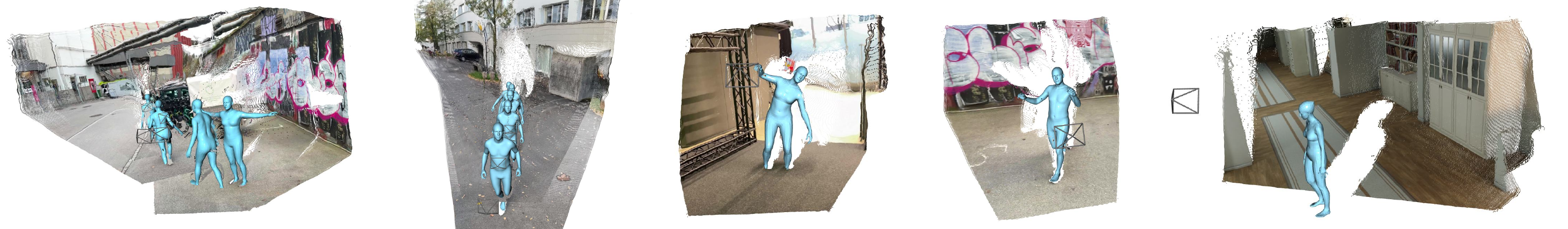}
\caption{
Qualitative results on joint human-scene reconstruction.
3R-Adapter produces more consistent human mesh recovery and cleaner scene geometry under complex human-scene interactions.
}
\label{fig-visualization-human}
\vspace{-10pt}
\end{figure}

\subsection{Analysis on Joint Human-Scene Reconstruction}

Joint human-scene reconstruction requires the model to go beyond standard point-level correspondence and capture structured associations between human proxy tokens, body regions, and surrounding scene context. 
This makes it a particularly suitable benchmark for evaluating whether relational adaptation can transfer a generic feed-forward reconstruction backbone to more structured downstream tasks.

As shown in Table~\ref{tab-humanscene}, 3R-Adapter achieves the strongest overall performance among PEFT baselines under the same CUT3R backbone, while using the fewest trainable parameters. 
This indicates that the improvement does not simply come from larger adaptation capacity, but from explicitly adapting relational structures through retrieval, rewiring, and refinement. 
Compared with channel-wise PEFT methods, our design is better aligned with the human-scene setting, where reconstruction depends on structured cross-entity relations rather than only local feature adjustment.

3R-Adapter also remains competitive with task-specific joint reconstruction methods, despite operating as a lightweight adaptation module for a generic reconstruction backbone. 
This suggests that relation-centric PEFT can effectively extend pre-trained 3D reconstruction models to structured downstream tasks without requiring a fully specialized architecture. 
The qualitative results in Fig.~\ref{fig-visualization-human} further support this observation: our method produces more coherent human meshes and more consistent scene reconstruction under challenging human-scene interactions, indicating that the adapted relational structures benefit both human and scene modeling.

\subsection{Quantitative Cross-view Correspondence Matching Accuracy}
\label{sec-correspondence-accuracy}
We further evaluate the cross-view correspondence matching accuracy of attended or retrieved reference locations with respect to geometry-consistent ground-truth correspondences.
Using $\pi^3$, we analyze 204 four-view inputs from 51 \textit{Oxford Day\&Night}~\cite{oxforddn} sequences, with correspondences obtained by 3D projection using ground-truth depth and camera geometry.
Each $518\times518$ image contains 1,369 patch tokens on a $37\times37$ grid, using $14\times14$-pixel patches.

Match@1 measures the fraction of valid query/reference pairs whose highest-attention token lies within a $w\times w$ patch window around the projected correspondence.
Window size $w$ denotes the geometric tolerance width in patch tokens; $w=1$ requires an exact token match.
These tolerances therefore retain local geometric constraints.
All reported metrics are percentages and adapted Match@1 and deformable offset coverage are averaged over three training seeds.

Table~\ref{tab-relational-diagnostics}(a) shows lower frozen Match@1 at night across all window sizes.
At $w=3$, it falls from 18.49\% by day to 11.14\% at night, confirming weaker cross-view matching.

Deformable offset coverage measures whether any reference location sampled by 3R-Adapter's learned offsets falls within the same window.
Table~\ref{tab-relational-diagnostics}(b) uses four candidates from one layer and 32 across eight layers per query/reference pair, then up to 64 across valid reference views per query.
At $w=3$, coverage is 28.72\%, 82.00\%, and 86.19\%, respectively.
Pooling candidates thus provides broader geometric support for rewiring; these coverage rates reflect different budgets and counting units and do not evaluate learned ranking accuracy.

Table~\ref{tab-relational-diagnostics}(c) further shows higher nighttime Match@1 after adaptation in all four reported blocks, using $w=3$.
Block 21 improves from 14.38\% to 19.21\%, and block 25 from 36.02\% to 44.41\%.
Block 31 changes from 52.94\% to 52.16\%, indicating that recovery is layer-dependent.
These results provide direct evidence that 3R-Adapter recovers geometric associations at intermediate decoder layers, consistent with our relation-centric design.

\begin{table*}[!t]
\centering
\caption{Cross-view correspondence diagnostics on \textit{Oxford Day\&Night} with $\pi^3$.
(a) Frozen Match@1 pooled over blocks 17/19/21/23.
(b) Coverage of 3R-Adapter's deformable retrieval offsets, pooled over day/night and evaluated within one layer, across all eight adapter layers, or across valid reference views.
(c) Paired nighttime Match@1 with a fixed $3\times3$ tolerance.
}
\label{tab-relational-diagnostics}
\vspace{5pt}
\resizebox{0.75\textwidth}{!}{%
\begin{minipage}{\textwidth}
\centering
\begin{minipage}[t]{0.23\linewidth}
\centering
\footnotesize
\textbf{(a) Frozen Match@1}\\[3pt]
\setlength{\tabcolsep}{1pt}
\renewcommand{\arraystretch}{1.1}
\begin{tabular*}{\linewidth}{@{\extracolsep{\fill}}ccc@{}}
\toprule
\textbf{Window size} & \textbf{Day} & \textbf{Night} \\
\midrule
1 & 7.56 & 3.69 \\
3 & 18.49 & 11.14 \\
5 & 24.38 & 16.64 \\
7 & 29.60 & 22.22 \\
\bottomrule
\end{tabular*}
\end{minipage}\hfill
\begin{minipage}[t]{0.49\linewidth}
\centering
\footnotesize
\textbf{(b) Deformable Offset Coverage}\\[3pt]
\setlength{\tabcolsep}{1pt}
\renewcommand{\arraystretch}{1.1}
\begin{tabular*}{\linewidth}{@{\extracolsep{\fill}}cccc@{}}
\toprule
\textbf{Window size} & \textbf{Single layer} & \textbf{All layers} & \textbf{All references} \\
\midrule
1 & 3.71 & 24.03 & 31.27 \\
3 & 28.72 & 82.00 & 86.19 \\
5 & 62.44 & 91.55 & 93.50 \\
7 & 83.26 & 94.74 & 96.05 \\
\bottomrule
\end{tabular*}
\end{minipage}\hfill
\begin{minipage}[t]{0.24\linewidth}
\centering
\footnotesize
\textbf{(c) Night Match@1}\\[3pt]
\setlength{\tabcolsep}{1pt}
\renewcommand{\arraystretch}{1.1}
\begin{tabular*}{\linewidth}{@{\extracolsep{\fill}}ccc@{}}
\toprule
\textbf{Block} & \textbf{Frozen} & \textbf{Adapted} \\
\midrule
19 & 2.86 & \cellcolor{gray!15}\textbf{3.85} \\
21 & 14.38 & \cellcolor{gray!15}\textbf{19.21} \\
23 & 26.47 & \cellcolor{gray!15}\textbf{33.19} \\
25 & 36.02 & \cellcolor{gray!15}\textbf{44.41} \\
\bottomrule
\end{tabular*}
\end{minipage}
\end{minipage}%
}
\end{table*}

\section{Conclusion}
\label{conclu}
We present 3R-Adapter, a relation-centric parameter-efficient adaptation framework for feed-forward 3D reconstruction models. 
Motivated by the observation that pre-trained models exhibit relational attention structure collapse in data-sparse long-tail scenarios, we reformulate adaptation as the recovery of cross-view relational structures rather than channel-wise feature adjustment. 
Our method retrieves task-relevant relational candidates, rewires them into structured cross-view associations, and refines predictions through consistency-preserving residual updates. 
Experiments on low-light reconstruction, joint human-scene reconstruction, and test-time adaptation demonstrate that 3R-Adapter improves robustness over existing PEFT methods with minimal trainable parameters. 
These results suggest that explicitly adapting relational structures is a promising direction for efficient and reliable adaptation of 3D reconstruction foundation models.

\begin{ack}
This work is supported by the National Natural Science Foundation of China (62376011) and the National Key R\&D Program of China (2024YFA1410000).
\end{ack}

\bibliographystyle{unsrt}
\bibliography{3RAdapter}

@inproceedings{cut3r,
  title={Continuous 3d perception model with persistent state},
  author={Wang, Qianqian and Zhang, Yifei and Holynski, Aleksander and Efros, Alexei A and Kanazawa, Angjoo},
  booktitle={Proceedings of the Computer Vision and Pattern Recognition Conference},
  pages={10510--10522},
  year={2025}
}

@article{human3r,
  title={Human3r: Everyone everywhere all at once},
  author={Chen, Yue and Chen, Xingyu and Xue, Yuxuan and Chen, Anpei and Xiu, Yuliang and Pons-Moll, Gerard},
  journal={arXiv preprint arXiv:2510.06219},
  year={2025}
}

@inproceedings{dust3r,
  title={Dust3r: Geometric 3d vision made easy},
  author={Wang, Shuzhe and Leroy, Vincent and Cabon, Yohann and Chidlovskii, Boris and Revaud, Jerome},
  booktitle={Proceedings of the IEEE/CVF conference on computer vision and pattern recognition},
  pages={20697--20709},
  year={2024}
}

@article{unish,
  title={UniSH: Unifying Scene and Human Reconstruction in a Feed-Forward Pass},
  author={Li, Mengfei and Li, Peng and Zhang, Zheng and Lu, Jiahao and Zhao, Chengfeng and Xue, Wei and Liu, Qifeng and Peng, Sida and Zhang, Wenxiao and Luo, Wenhan and others},
  journal={arXiv preprint arXiv:2601.01222},
  year={2026}
}

@article{josh3r,
  title={Joint optimization for 4d human-scene reconstruction in the wild},
  author={Liu, Zhizheng and Lin, Joe and Wu, Wayne and Zhou, Bolei},
  journal={arXiv preprint arXiv:2501.02158},
  year={2025}
}

@article{croco,
  title={Croco: Self-supervised pre-training for 3d vision tasks by cross-view completion},
  author={Weinzaepfel, Philippe and Leroy, Vincent and Lucas, Thomas and Br{\'e}gier, Romain and Cabon, Yohann and Arora, Vaibhav and Antsfeld, Leonid and Chidlovskii, Boris and Csurka, Gabriela and Revaud, J{\'e}r{\^o}me},
  journal={Advances in Neural Information Processing Systems},
  volume={35},
  pages={3502--3516},
  year={2022}
}

@inproceedings{mast3r,
  title={Grounding image matching in 3d with mast3r},
  author={Leroy, Vincent and Cabon, Yohann and Revaud, J{\'e}r{\^o}me},
  booktitle={European conference on computer vision},
  pages={71--91},
  year={2024},
  organization={Springer}
}

@inproceedings{spann3r,
  title={3d reconstruction with spatial memory},
  author={Wang, Hengyi and Agapito, Lourdes},
  booktitle={2025 International Conference on 3D Vision (3DV)},
  pages={78--89},
  year={2025},
  organization={IEEE}
}

@article{online3r,
  title={Online3R: Online Learning for Consistent Sequential Reconstruction Based on Geometry Foundation Model},
  author={Zhou, Shunkai and Yan, Zike and Xue, Fei and Wu, Dong and Deng, Yuchen and Zha, Hongbin},
  journal={arXiv preprint arXiv:2604.09480},
  year={2026}
}

@article{oxforddn,
  title={Seeing in the dark: Benchmarking egocentric 3d vision with the oxford day-and-night dataset},
  author={Wang, Zirui and Bian, Wenjing and Li, Xinghui and Tao, Yifu and Wang, Jianeng and Fallon, Maurice and Prisacariu, Victor Adrian},
  journal={arXiv preprint arXiv:2506.04224},
  year={2025}
}

@article{dark3r,
  title={Dark3R: Learning Structure from Motion in the Dark},
  author={Guo, Andrew Y and Malik, Anagh and Tedla, SaiKiran and Dai, Yutong and Qin, Yiqian and Salehe, Zach and Attal, Benjamin and Nousias, Sotiris and Kutulakos, Kiriakos N and Lindell, David B},
  journal={arXiv preprint arXiv:2603.05330},
  year={2026}
}

@inproceedings{dora,
  title={Dora: Weight-decomposed low-rank adaptation},
  author={Liu, Shih-Yang and Wang, Chien-Yi and Yin, Hongxu and Molchanov, Pavlo and Wang, Yu-Chiang Frank and Cheng, Kwang-Ting and Chen, Min-Hung},
  booktitle={Forty-first International Conference on Machine Learning},
  year={2024}
}

@article{dept,
  title={Dept: Decomposed prompt tuning for parameter-efficient fine-tuning},
  author={Shi, Zhengxiang and Lipani, Aldo},
  journal={arXiv preprint arXiv:2309.05173},
  year={2023}
}

@inproceedings{nightadapter,
  title={NightAdapter: Learning a frequency adapter for generalizable night-time scene segmentation},
  author={Bi, Qi and Yi, Jingjun and Huang, Huimin and Zheng, Hao and Zhan, Haolan and Huang, Yawen and Li, Yuexiang and Wu, Xian and Zheng, Yefeng},
  booktitle={Proceedings of the Computer Vision and Pattern Recognition Conference},
  pages={23838--23849},
  year={2025}
}

@inproceedings{hydralora,
  title={HydraLoRA: An Asymmetric LoRA Architecture for Efficient Fine-Tuning},
  author={Tian, Chunlin and Shi, Zhan and Guo, Zhijiang and Li, Li and Xu, Chengzhong},
  booktitle={Advances in Neural Information Processing Systems (NeurIPS)},
  year={2024}
}

@inproceedings{tlora,
  title={T-lora: Single image diffusion model customization without overfitting},
  author={Soboleva, Vera and Alanov, Aibek and Kuznetsov, Andrey and Sobolev, Konstantin},
  booktitle={Proceedings of the AAAI Conference on Artificial Intelligence},

  pages={9051--9059},
  year={2026}
}

@article{pi3,
  title={$\pi^3$: Permutation-Equivariant Visual Geometry Learning},
  author={Wang, Yifan and Zhou, Jianjun and Zhu, Haoyi and Chang, Wenzheng and Zhou, Yang and Li, Zizun and Chen, Junyi and Pang, Jiangmiao and Shen, Chunhua and He, Tong},
  journal={arXiv preprint arXiv:2507.13347},
  year={2025}
}

@inproceedings{multihmr,
  title={Multi-hmr: Multi-person whole-body human mesh recovery in a single shot},
  author={Baradel, Fabien and Armando, Matthieu and Galaaoui, Salma and Br{\'e}gier, Romain and Weinzaepfel, Philippe and Rogez, Gr{\'e}gory and Lucas, Thomas},
  booktitle={European Conference on Computer Vision},
  pages={202--218},
  year={2024},
  organization={Springer}
}

@inproceedings{vggt,
  title={Vggt: Visual geometry grounded transformer},
  author={Wang, Jianyuan and Chen, Minghao and Karaev, Nikita and Vedaldi, Andrea and Rupprecht, Christian and Novotny, David},
  booktitle={Proceedings of the Computer Vision and Pattern Recognition Conference},
  pages={5294--5306},
  year={2025}
}

@misc{test3r,
  title={Test3R: Learning to Reconstruct 3D at Test Time}, 
  author={Yuheng Yuan and Qiuhong Shen and Shizun Wang and Xingyi Yang and Xinchao Wang},
  year={2025},
  eprint={2506.13750},
  archivePrefix={arXiv},
  primaryClass={cs.CV},
  url={https://arxiv.org/abs/2506.13750}, 
}

@inproceedings{gaprompt,
  title = 	 {{GAP}rompt: Geometry-Aware Point Cloud Prompt for 3{D} Vision Model},
  author =    {Ai, Zixiang and Liu, Zichen and Lei, Yuanhang and Cui, Zhenyu and Zou, Xu and Zhou, Jiahuan},
  booktitle = {Proceedings of the 42nd International Conference on Machine Learning},
  pages = 	 {796--808},
  year = 	 {2025},
  volume = 	 {267},
  series = 	 {Proceedings of Machine Learning Research},
  month = 	 {13--19 Jul},
  publisher = {PMLR}
}

@article{ai2026gaprompt++,
  title={{GAPrompt++}: Multi-Granular Geometry-Aware Point Cloud Prompt for {3D} Vision Model},
  author={Ai, Zixiang and Cui, Zhenyu and Guo, Yufei and Qiang, Wenwen and Chen, Lei and Lu, Jiwen and Zhou, Jiahuan},
  journal={IEEE Transactions on Pattern Analysis and Machine Intelligence},
  year={2026},
  publisher={IEEE}
}

@inproceedings{7scene,
  title={Scene coordinate regression forests for camera relocalization in RGB-D images},
  author={Shotton, Jamie and Glocker, Ben and Zach, Christopher and Izadi, Shahram and Criminisi, Antonio and Fitzgibbon, Andrew},
  booktitle={Proceedings of the IEEE conference on computer vision and pattern recognition},
  pages={2930--2937},
  year={2013}
}

@inproceedings{nrgbd,
  title={Neural rgb-d surface reconstruction},
  author={Azinovi{\'c}, Dejan and Martin-Brualla, Ricardo and Goldman, Dan B and Nie{\ss}ner, Matthias and Thies, Justus},
  booktitle={Proceedings of the IEEE/CVF conference on computer vision and pattern recognition},
  pages={6290--6301},
  year={2022}
}

@inproceedings{bedlam,
  title={Bedlam: A synthetic dataset of bodies exhibiting detailed lifelike animated motion},
  author={Black, Michael J and Patel, Priyanka and Tesch, Joachim and Yang, Jinlong},
  booktitle={Proceedings of the IEEE/CVF Conference on Computer Vision and Pattern Recognition},
  pages={8726--8737},
  year={2023}
}

@article{da3,
  title={Depth anything 3: Recovering the visual space from any views},
  author={Lin, Haotong and Chen, Sili and Liew, Junhao and Chen, Donny Y and Li, Zhenyu and Shi, Guang and Feng, Jiashi and Kang, Bingyi},
  journal={arXiv preprint arXiv:2511.10647},
  year={2025}
}

@inproceedings{scannet,
  title={Scannet: Richly-annotated 3d reconstructions of indoor scenes},
  author={Dai, Angela and Chang, Angel X and Savva, Manolis and Halber, Maciej and Funkhouser, Thomas and Nie{\ss}ner, Matthias},
  booktitle={Proceedings of the IEEE conference on computer vision and pattern recognition},
  pages={5828--5839},
  year={2017}
}

@inproceedings{dl3dv,
  title={Dl3dv-10k: A large-scale scene dataset for deep learning-based 3d vision},
  author={Ling, Lu and Sheng, Yichen and Tu, Zhi and Zhao, Wentian and Xin, Cheng and Wan, Kun and Yu, Lantao and Guo, Qianyu and Yu, Zixun and Lu, Yawen and others},
  booktitle={Proceedings of the IEEE/CVF Conference on Computer Vision and Pattern Recognition},
  pages={22160--22169},
  year={2024}
}

@inproceedings{wildrgbd,
  title={Rgbd objects in the wild: Scaling real-world 3d object learning from rgb-d videos},
  author={Xia, Hongchi and Fu, Yang and Liu, Sifei and Wang, Xiaolong},
  booktitle={Proceedings of the IEEE/CVF Conference on Computer Vision and Pattern Recognition},
  pages={22378--22389},
  year={2024}
}

@inproceedings{megadepth,
  title={Megadepth: Learning single-view depth prediction from internet photos},
  author={Li, Zhengqi and Snavely, Noah},
  booktitle={Proceedings of the IEEE conference on computer vision and pattern recognition},
  pages={2041--2050},
  year={2018}
}

@inproceedings{co3d,
  title={Common objects in 3d: Large-scale learning and evaluation of real-life 3d category reconstruction},
  author={Reizenstein, Jeremy and Shapovalov, Roman and Henzler, Philipp and Sbordone, Luca and Labatut, Patrick and Novotny, David},
  booktitle={Proceedings of the IEEE/CVF international conference on computer vision},
  pages={10901--10911},
  year={2021}
}

@inproceedings{tartanair,
  title={Tartanair: A dataset to push the limits of visual slam},
  author={Wang, Wenshan and Zhu, Delong and Wang, Xiangwei and Hu, Yaoyu and Qiu, Yuheng and Wang, Chen and Hu, Yafei and Kapoor, Ashish and Scherer, Sebastian},
  booktitle={2020 IEEE/RSJ International Conference on Intelligent Robots and Systems (IROS)},
  pages={4909--4916},
  year={2020},
  organization={IEEE}
}

@inproceedings{mvimgnet,
    title     = {MVImgNet: A Large-scale Dataset of Multi-view Images},
    author    = {Yu, Xianggang and Xu, Mutian and Zhang, Yidan and Liu, Haolin and Ye, Chongjie and Wu, Yushuang and Yan, Zizheng and Liang, Tianyou and Chen, Guanying and Cui, Shuguang and Han, Xiaoguang},
    booktitle = {CVPR},
    year      = {2023}
}

@article{lora,
  title={Lora: Low-rank adaptation of large language models.},
  author={Hu, Edward J and Shen, Yelong and Wallis, Phillip and Allen-Zhu, Zeyuan and Li, Yuanzhi and Wang, Shean and Wang, Liang and Chen, Weizhu and others},
  journal={Iclr},
  volume={1},
  number={2},
  pages={3},
  year={2022}
}

@inproceedings{ai2025vgp,
  title = 	 {Vision Graph Prompting via Semantic Low-Rank Decomposition},
  author =   {Ai, Zixiang and Liu, Zichen and Zhou, Jiahuan},
  booktitle = {Proceedings of the 42nd International Conference on Machine Learning},
  pages = 	 {809--821},
  year = 	 {2025},
  volume = 	 {267},
  series = 	 {Proceedings of Machine Learning Research},
  month = 	 {13--19 Jul},
  publisher = {PMLR}
}

@inproceedings{sfm,
  title={Structure-from-motion revisited},
  author={Schonberger, Johannes L and Frahm, Jan-Michael},
  booktitle={Proceedings of the IEEE conference on computer vision and pattern recognition},
  pages={4104--4113},
  year={2016}
}

@inproceedings{mvs,
  title={Multi-view stereo revisited},
  author={Goesele, Michael and Curless, Brian and Seitz, Steven M},
  booktitle={2006 IEEE Computer Society Conference on Computer Vision and Pattern Recognition (CVPR'06)},
  volume={2},
  pages={2402--2409},
  year={2006},
  organization={IEEE}
}

@inproceedings{vpt,
  title={Visual prompt tuning},
  author={Jia, Menglin and Tang, Luming and Chen, Bor-Chun and Cardie, Claire and Belongie, Serge and Hariharan, Bharath and Lim, Ser-Nam},
  booktitle={European Conference on Computer Vision},
  pages={709--727},
  year={2022},
  organization={Springer}
}

@inproceedings{adapter,
  title={Parameter-efficient transfer learning for NLP},
  author={Houlsby, Neil and Giurgiu, Andrei and Jastrzebski, Stanislaw and Morrone, Bruna and De Laroussilhe, Quentin and Gesmundo, Andrea and Attariyan, Mona and Gelly, Sylvain},
  booktitle={International conference on machine learning},
  pages={2790--2799},
  year={2019},
  organization={PMLR}
}

@article{gen3r,
  title={Gen3R: 3D Scene Generation Meets Feed-Forward Reconstruction},
  author={Huang, Jiaxin and Yang, Yuanbo and Yang, Bangbang and Ma, Lin and Ma, Yuewen and Liao, Yiyi},
  journal={arXiv preprint arXiv:2601.04090},
  year={2026}
}

@article{vist3a,
  title={VIST3A: Text-to-3D by Stitching a Multi-view Reconstruction Network to a Video Generator},
  author={Go, Hyojun and Narnhofer, Dominik and Bhat, Goutam and Truong, Prune and Tombari, Federico and Schindler, Konrad},
  journal={arXiv preprint arXiv:2510.13454},
  year={2025}
}

@article{vggrpo,
  title={VGGRPO: Towards World-Consistent Video Generation with 4D Latent Reward},
  author={An, Zhaochong and Kupyn, Orest and Uscidda, Th{\'e}o and Colaco, Andrea and Ahuja, Karan and Belongie, Serge and Gonzalez-Franco, Mar and Gazulla, Marta Tintore},
  journal={arXiv preprint arXiv:2603.26599},
  year={2026}
}

@article{smpl,
      author = {Loper, Matthew and Mahmood, Naureen and Romero, Javier and Pons-Moll, Gerard and Black, Michael J.},
      title = {{SMPL}: A Skinned Multi-Person Linear Model},
      journal = {ACM Trans. Graphics (Proc. SIGGRAPH Asia)},
      month = oct,
      number = {6},
      pages = {248:1--248:16},
      publisher = {ACM},
      volume = {34},
      year = {2015}
    }



\appendix
\newpage
\section{Technical appendices and supplementary material}


\subsection{Analysis and Discussion}

Dense attention offers a flexible way to learn cross-view relations during large-scale pre-training, but it also introduces an unconstrained all-to-all relational search space. 
In data-sparse long-tail scenarios, directly adapting such dense relations is inefficient and may amplify noisy correspondences. 
Our Deformable Retrieval can be interpreted as a structured sparse approximation to dense cross-view attention. 
The sliding window mask provides a weak view-locality prior, while learnable offsets retain spatial flexibility for task-adaptive correspondence search. 
This design reduces the difficulty of relational discovery without imposing hard geometric constraints.

This perspective also distinguishes 3R-Adapter from conventional PEFT methods. 
Rather than using trainable parameters only for channel-wise feature adjustment, our method uses them to adapt how cross-view relations are retrieved, rewired, and injected into predictions. 
Thus, the parameter efficiency of 3R-Adapter comes not from a larger adaptation capacity, but from placing lightweight trainable modules at the relational bottleneck of feed-forward 3D reconstruction.

\subsection{Limitation}

\label{limit}

Although 3R-Adapter effectively alleviates the degradation of feed-forward reconstruction models in data-sparse long-tail scenarios, it still relies on the representational and geometric priors learned by the pre-trained backbone. 
Our method adapts and reorganizes existing cross-view relational structures rather than learning reconstruction ability from scratch. 
Thus, when the backbone fails to extract meaningful visual features or the input distribution lies far outside the pre-training regime, relational retrieval and rewiring may remain limited. 
In addition, the current sliding-window design improves efficiency but may restrict long-range relation modeling in extremely long sequences or scenes with large viewpoint changes. 
Future work may explore adaptive view selection and hierarchical relational memory for more scalable adaptation.

\subsection{Details of Datasets and Metrics}

Due to the scarcity of high-quality annotations in long-tail scenarios, we evaluate our method on representative benchmarks that reflect diverse forms of domain shift. 
We report task-specific metrics for camera pose estimation, scene reconstruction, and human mesh reconstruction. 
Unless otherwise specified, camera trajectory and scene reconstruction errors are measured in meters, while human reconstruction errors are measured in millimeters.

\textbf{Low-light reconstruction.}
We use the \textit{Oxford Day\&Night}~\cite{oxforddn} dataset, which provides calibrated camera poses and semi-dense point clouds captured under varying illumination conditions, including day, dusk, and night. 
This dataset enables systematic evaluation of reconstruction robustness under severe appearance degradation. 
For camera pose estimation, we report Absolute Trajectory Error (ATE), Relative Pose Error in translation (RPE-T), and Relative Pose Error in rotation (RPE-R). 
ATE measures the global trajectory discrepancy after alignment, RPE-T measures relative translation drift between frames, and RPE-R measures relative rotational error. 
ATE and RPE-T are reported in meters, while RPE-R is reported in degrees.

For scene reconstruction, we report Accuracy, Completeness, Chamfer Distance, and Normal Consistency. 
Accuracy measures the average distance from predicted points to the ground-truth point cloud, while Completeness measures the average distance from ground-truth points to the prediction. 
Chamfer Distance is computed as the sum of the two directional distances:
\begin{equation}
    \mathrm{CD}(\mathcal{P}, \mathcal{G})
    =
    \frac{1}{|\mathcal{P}|}\sum_{\mathbf{p}\in \mathcal{P}}
    \min_{\mathbf{g}\in \mathcal{G}}\|\mathbf{p}-\mathbf{g}\|_2
    +
    \frac{1}{|\mathcal{G}|}\sum_{\mathbf{g}\in \mathcal{G}}
    \min_{\mathbf{p}\in \mathcal{P}}\|\mathbf{g}-\mathbf{p}\|_2 ,
\end{equation}
where $\mathcal{P}$ and $\mathcal{G}$ denote the predicted and ground-truth point sets. 
Accuracy, Completeness, and Chamfer Distance are reported in meters. 
Normal Consistency measures the cosine similarity between predicted and ground-truth surface normals, where higher values indicate better local geometric consistency.

\textbf{Joint human-scene reconstruction.}
We adopt the \textit{BEDLAM}~\cite{bedlam} dataset, which provides both 3D scene geometry and human SMPL parameters as ground truth. 
The dataset is synthetically generated across diverse indoor and outdoor environments, allowing controlled evaluation of human-scene interaction. 

For scene reconstruction, we use the same Accuracy, Completeness, Chamfer Distance, and Normal Consistency metrics as described above. 
For human mesh reconstruction, we report PA-PVE, Metric-PVE, PA-MPJPE, and Metric-MPJPE. 
PVE measures the mean per-vertex error between predicted and ground-truth human meshes, while MPJPE measures the mean per-joint position error. 
The prefix PA indicates evaluation after Procrustes alignment, which removes global similarity transformation differences. 
In contrast, Metric denotes evaluation in the original metric space without alignment, and therefore reflects absolute human localization and the spatial consistency between the reconstructed human body and the scene. 
All human reconstruction metrics are reported in millimeters.

\textbf{Test-time adaptation.}
Following the Test3R/CUT3R sparse-view protocol, we sample every 200 frames on \textit{7-Scenes}~\cite{7scene} and every 500 frames on \textit{NRGBD}~\cite{nrgbd} to simulate test-time adaptation scenarios.
Each sequence therefore contains approximately 3--5 views on \textit{7-Scenes} and 2--4 views on \textit{NRGBD}, requiring the model to reconstruct geometry under self-supervised conditions without additional training data.
We evaluate scene reconstruction using Accuracy, Completeness, Chamfer Distance, and Normal Consistency, with distance-based metrics reported in meters.

\textbf{Cross-view attention diagnostics.}
GT-valid query/reference pairs are determined by projection and reprojection with dataset depth and camera geometry, and shared by all models.
$H_{\mathrm{n}}=-\sum_j p_j\log p_j/\log N$ uses the conditional reference distribution over $N$ patch tokens.
The frozen backbone is evaluated once; adapted statistics use three released checkpoints (seeds 1111/1112/1113).
All confidence intervals use 10,000 whole-sequence bootstrap replicates stratified by site and day/night, preserving input, layer, and model pairing.
They describe sequence sampling within the five observed sites, conditional on the three checkpoints, rather than training-seed population uncertainty.
Passive measurement leaves model outputs unchanged.

\begin{table}[h] 
\centering
\caption{Ablation of components in 3R-Adapter on the \textit{Oxford Day\&Night}~\cite{oxforddn} test dataset for low-light reconstruction.}
\begin{scriptsize}
\resizebox{0.9\linewidth}{!}{
\begin{tabular}{ccc|cccc}
\toprule 
\textbf{Deformable Retrieval}&\textbf{Relational Rewiring}&\textbf{Consistency Refinement}& \textbf{Accuracy} $\downarrow$ 
& \textbf{Complete} $\downarrow$ 
& \textbf{Chamfer} $\downarrow$ 
& \textbf{Normal} $\uparrow$ \\ 
\midrule
- & - & - & 0.562 & 0.639 & 1.201 & 0.534 \\ 
\checkmark & - & - & 0.540 & 0.612 & 1.152 & 0.534 \\ 
\checkmark & \checkmark & - & 0.397 & 0.445 & 0.842 & 0.542 \\ 
- & - &\checkmark & 0.483 & 0.529 & 1.012 & 0.536 \\ 
\checkmark & \checkmark & \checkmark & \textbf{0.254} & \textbf{0.327} & \textbf{0.581} & \textbf{0.550}  \\    
\bottomrule
\end{tabular}
}
\end{scriptsize}
\label{tab-ablation-components}
\vspace{-10pt}
\end{table}

\subsection{Detailed Ablation Study}

\paragraph{Ablation on main components.}Table~\ref{tab-ablation-components} studies the contribution of each component in 3R-Adapter on the Oxford Day\&Night dataset using CUT3R as the backbone. 
The vanilla CUT3R baseline obtains a Chamfer Distance of 1.201, indicating substantial degradation under low-light conditions.

Adding Deformable Retrieval alone slightly improves the reconstruction results, reducing Chamfer distance from 1.201 to 1.152. 
This suggests that sparse candidate querying can reduce noisy dense interactions, but retrieving candidate locations alone is insufficient to recover stable cross-view structure. 
When Relational Rewiring is further introduced, performance improves significantly, with Chamfer distance reduced to 0.842 and Accuracy improved from 0.540 to 0.397. 
This confirms that the key benefit comes from explicitly reorganizing retrieved candidates into structured cross-view relations.

Consistency Refinement also provides clear gains when used alone, improving Chamfer distance from 1.201 to 1.012. 
This shows that lightweight residual correction can stabilize predictions even without explicit retrieval and rewiring. 
However, the full 3R-Adapter achieves the best results across all metrics, reducing Chamfer distance to \textbf{0.581} and improving Normal consistency to \textbf{0.550}. 
These results demonstrate the complementary roles of the three components: Deformable Retrieval identifies reliable candidates, Relational Rewiring reconstructs cross-view associations, and Consistency Refinement safely injects the adapted relations into the prediction stream.

\paragraph{Ablation on Hyperparameters.}
We further analyze the sensitivity of 3R-Adapter to four key hyperparameters: the number of sampling locations $K$, the search radius $\rho$, the relational fusion factor $\alpha$, and the residual scaling factor $\gamma$. We report Chamfer Distance on the \textit{Oxford Day\&Night} dataset, where lower values indicate better reconstruction quality.

\begin{figure}[!t]
\centering
\includegraphics[width=\textwidth]{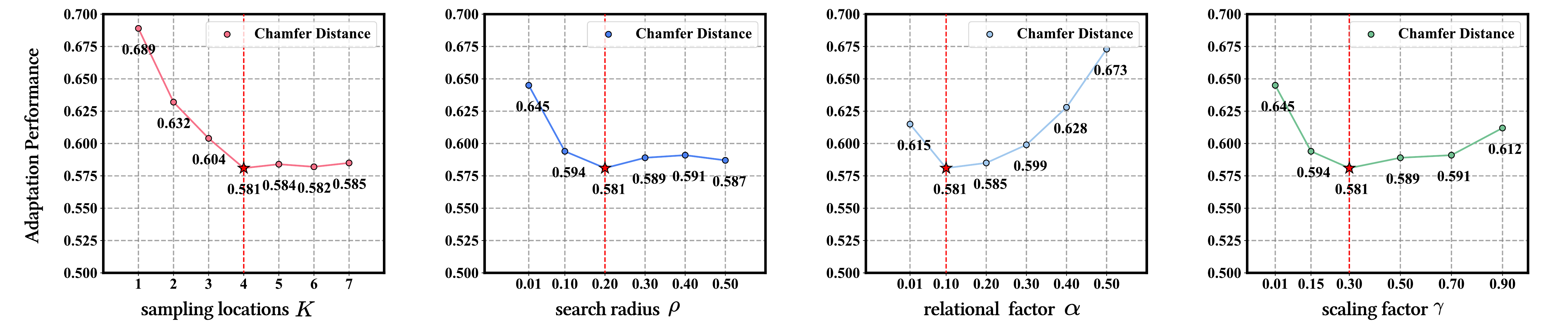}
\caption{
Ablation on the hyperparameters of 3R-Adapter on the \textit{Oxford Day\&Night}~\cite{oxforddn} test dataset for scene reconstruction.
}
\label{fig-visualization-hyperparam}
\end{figure}

As shown in Fig.~\ref{fig-visualization-hyperparam}, increasing the number of sampling locations from $K=1$ to $K=4$ consistently improves performance, suggesting that multiple candidates provide richer relational evidence for cross-view reconstruction. However, using more than four samples brings marginal gains and may introduce redundant or noisy candidates. The search radius $\rho$ exhibits a similar trend. A moderate radius improves local relation discovery, with the best performance achieved at $\rho=0.20$, while an overly large radius tends to include irrelevant distant features.

The relational fusion factor $\alpha$ controls the contribution of rewired cross-view relations. The model performs best at $\alpha=0.10$, while larger values degrade performance, indicating that excessive relational injection can disturb the pre-trained feature distribution. The residual scaling factor $\gamma$ shows a clear balance between adaptation strength and prior preservation, with the best result obtained at $\gamma=0.30$.

Overall, 3R-Adapter achieves the best configuration with $K=4$, $\rho=0.20$, $\alpha=0.10$, and $\gamma=0.30$, which is used as the default setting in our experiments.

\subsection{Implementation Details}
\label{impl}
We build 3R-Adapter on top of pretrained feed-forward 3D reconstruction backbones, including DUSt3R~\cite{dust3r}, CUT3R~\cite{cut3r}, and $\pi^3$~\cite{pi3}. 
Across all experiments, the pretrained backbone is frozen, and only the lightweight adapter parameters are updated. 
This setting allows the model to preserve geometric priors learned from large-scale pre-training while adapting to data-sparse long-tail scenarios with minimal trainable parameters.
We optimize the adapter modules using AdamW. 
The hyperparameters of 3R-Adapter are set according to the ablation study: the number of deformable sampling points is $K=4$, the search radius is $\rho=0.20$, the relational fusion factor is $\alpha=0.10$, and the residual scaling factor is $\gamma=0.30$. 
All experiments are conducted on 8 NVIDIA RTX 5090 GPUs.
For low-light reconstruction and joint human-scene reconstruction, we train the adapters on the corresponding training splits and report results on the held-out evaluation sets. 
For test-time adaptation, the adapter is optimized independently on each test scene using self-supervised reconstruction loss~\cite{test3r}, without accessing additional training data or ground-truth geometry.

\subsection{Training Objectives}

We optimize 3R-Adapter using the standard objective of each adaptation task, while keeping the pre-trained backbone frozen. 
This ensures that the comparison focuses on the adaptation mechanism rather than additional supervision.

For low-light reconstruction, we use geometric reconstruction losses on predicted point maps or depth representations, together with pose supervision when camera trajectories are available:
\begin{equation}
    \mathcal{L}_{\text{low}} =
    \mathcal{L}_{\text{geo}} + \lambda_{\text{pose}}\mathcal{L}_{\text{pose}}.
\end{equation}

For joint human-scene reconstruction, we jointly supervise scene geometry and human body reconstruction:
\begin{equation}
    \mathcal{L}_{\text{hs}}
    =
    \mathcal{L}_{\text{scene}}
    +
    \lambda_{\text{mesh}}\mathcal{L}_{\text{mesh}}
    +
    \lambda_{\text{joint}}\mathcal{L}_{\text{joint}}
    +
    \lambda_{\text{smpl}}\mathcal{L}_{\text{smpl}}.
\end{equation}
Following Human3R~\cite{human3r}, we incorporate human prior tokens from Multi-HMR~\cite{multihmr} and fuse them with reconstruction tokens to enhance human-aware relational modeling. 
These tokens provide body-structure cues that help the model associate human proxy representations with body regions, while 3R-Adapter adapts the cross-view relational structures of the reconstruction backbone.

For test-time adaptation, we follow the self-supervised objective of Test3R~\cite{test3r}. 
Given sparse input views, the model predicts pointmaps under different reference-view choices. 
Since these predictions correspond to the same underlying scene geometry, we enforce consistency among pointmaps generated from different references:
\begin{equation}
    \mathcal{L}_{\text{tta}}
    =
    \sum_{r_1 \neq r_2}
    \left\|
    \mathbf{P}^{(r_1)} - \mathbf{P}^{(r_2)}
    \right\|_1,
\end{equation}
where $\mathbf{P}^{(r)}$ denotes the predicted pointmap when view $r$ is selected as the reference view. 
This self-supervised loss enables scene-level adaptation without ground-truth geometry.

\subsection{Broader Impact}
\label{impact}
3R-Adapter aims to improve the adaptability and robustness of feed-forward 3D reconstruction models in data-sparse scenarios, such as low-light environments, joint human-scene reconstruction, and test-time adaptation. By reducing the number of trainable parameters required for downstream adaptation, our method may lower the computational and data costs of deploying 3D reconstruction systems in practical settings, including robotics, augmented reality, digital content creation, and embodied intelligence.

At the same time, improved 3D reconstruction capabilities may raise privacy and safety concerns when applied to real-world environments or human-centered scenes. In particular, reconstructing human geometry or private spaces from visual inputs should be conducted with appropriate consent, data protection, and usage constraints. Our work focuses on technical improvements for efficient adaptation and does not introduce new data collection procedures. We encourage future applications of such systems to follow responsible data governance practices and to carefully consider potential misuse in surveillance or unauthorized 3D reconstruction.

\end{document}